\documentclass[runningheads,a4paper]{llncs}

\usepackage{amssymb}
\usepackage{graphicx}
\usepackage{subfigure}
\usepackage[cmex10]{amsmath}
\usepackage{color}
\usepackage{algorithmic}
\usepackage{algorithm}
\usepackage{multirow}
\usepackage[misc]{ifsym}
\usepackage{ifsym}
\usepackage{amsfonts}
\usepackage{array}
\usepackage{mathrsfs}

\newcolumntype{C}[1]{>{\centering}p{#1}}
\newcolumntype{L}[1]{>{\raggedleft}p{#1}}
\newcolumntype{R}[1]{>{\raggedright}p{#1}}
\usepackage{url}

\urldef{\mailsa}\path|xy_jia@foxmail.com,|
\urldef{\mailsb}\path|xmxu@scut.edu.cn,|
\urldef{\mailsc}\path|{caibolun,eecollinguo}@gmail.com|
\newcommand{\keywords}[1]{\par\addvspace\baselineskip
\noindent\keywordname\enspace\ignorespaces#1}

\begin{document}

\mainmatter  

\title{Single Image Super-Resolution Using \\ Multi-Scale Convolutional Neural Network}

\titlerunning{Single Image Super-Resolution Using Multi-Scale CNN}

%
%
\author{Xiaoyi Jia \and Xiangmin Xu\textsuperscript{(\Letter)}
\thanks{This work is supported by the National Natural Science Foundation of China (61171142, 61401163, U1636218), the Science and technology Planning Project of Guangdong Province of China (2014B010111003, 2014B010111006), Guangzhou Key Lab of Body Data Science (201605030011).
}
\and Bolun Cai \and Kailing Guo
}
\authorrunning{X. Jia et al.}

\institute{School of Electronic and Information Engineering\\
South China University of Technology, Guangzhou, China\\
\mailsa \mailsb \mailsc}

\maketitle

\begin{abstract}

Methods based on convolutional neural network (CNN) have demonstrated tremendous improvements on single image super-resolution. However, the previous methods mainly restore images from one single area in the low resolution (LR) input, which limits the flexibility of models to infer various scales of details for high resolution (HR) output. Moreover, most of them train a specific model for each up-scale factor. In this paper, we propose a multi-scale super resolution (MSSR) network. Our network consists of multi-scale paths to make the HR inference, which can learn to synthesize features from different scales. This property helps reconstruct various kinds of regions in HR images. In addition, only one single model is needed for multiple up-scale factors, which is more efficient without loss of restoration quality. Experiments on four public datasets demonstrate that the proposed method achieved state-of-the-art performance with fast speed.
\keywords{Super-resolution, convolutional neural network, multi-scale}
\end{abstract}

\section{Introduction}

The task of single image super-resolution aims at restoring a high-resolution (HR) image from a given low-resolution (LR) one. Super-resolution has wide applications in many fields where image details are on demand, such as medical, remote sensing imaging, video surveillance, and entertainment. In the past decades, super-resolution has attracted much attention from computer vision communities. Early methods include bicubic interpolation \cite{bicubic}, Lanczos resampling \cite{lanczos}, statistical priors \cite{prior}, neighbor embedding \cite{embed}, and sparse coding \cite{srsr}. However, super-resolution is highly ill-posed since the process from HR to LR contains non-invertible operation such as low-pass filtering and subsampling.

Deep convolutional neural networks (CNNs) have achieved state-of-the-art performance in computer vision, such as image classification \cite{vgg}, object detection \cite{rcnn}, and image enhancement \cite{dehazenet}. Recently, CNNs are widely used to address the ill-posed inverse problem of super-resolution, and have
demonstrated superiority over traditional methods \cite{lanczos,prior,embed,srsr} with respect to both reconstruction accuracy and computational efficiency. Dong et al. \cite{srcnn1,srcnn2} successfully design a super-resolution convolutional neural network (SRCNN) to demonstrate that a CNN can be applied to learn the mapping from LR to HR in an end-to-end manner. A fast super-resolution convolutional neural network (FSRCNN) \cite{fsrcnn} is proposed to accelerate the speed of SRCNN \cite{srcnn1,srcnn2}, which takes the original LR image as input and adopts a deconvolution layer to replace the bicubic interpolation. In \cite{subpixel}, an efficient sub-pixel convolution layer is introduced to achieve real time performance. Kim et al. \cite{vdsr} uses a very deep super-resolution (VDSR) network with 20 convolutional layers, which greatly improves the accuracy of the model.

The previous methods based on CNN has achieved great progress on the restoration quality as well as efficiency. However, there are some limitations mainly coming from the following aspects:
\begin{itemize}
  \item CNN based methods make efforts to enlarge the receptive field of the models as well as stack more layers. They reconstruct any type of contents from LR images using only single-scale region, thus ignore the various scales of different details. For instance, restoring the detail in the sky probably relies on a lager image region, while the tiny text may only be relevant to a small patch.
  \item Most previous approaches learn a specific model for one single up-scale factor. Therefore, the model learned for one up-scale factor cannot work well for another. That is, many scale-specific models should be trained for different up-scale factors, which is inefficient both in terms of time and memory. Though \cite{vdsr} trains a model for multiple up-scales, it ignores the fact that a single receptive field may contain different information amount in various resolution versions.
\end{itemize}

In this paper, we propose a multi-scale super resolution (MSSR) convolutional neural network to issue these problems -- there are two folds of meaning in the term multi-scale. First, the proposed network combines multi-path subnetworks with different depth, which correspond to multi-scale regions in the input image. Second, the multi-scale network is capable to select a proper receptive field for different up-scales to restore the HR image. Only single model is trained for multiple up-scale factors by multi-scale training. 

\section{Mutli-scale Super-Resolution}

Given a low-resolution image, super-resolution aims at restoring its high-resolution version. For this ill-posed recovery problem, it is probably an effective way to estimate a target pixel by taking into account more context information in the neighborhood. In \cite{srcnn1,srcnn2,vdsr}, authors found that larger receptive field tends to achieve better performance due to richer structural information. However, we argue that the restoration process is not only depending on single-scale regions with large receptive field.

\begin{figure}[ht]
  \centering
  \includegraphics[width=1\linewidth]{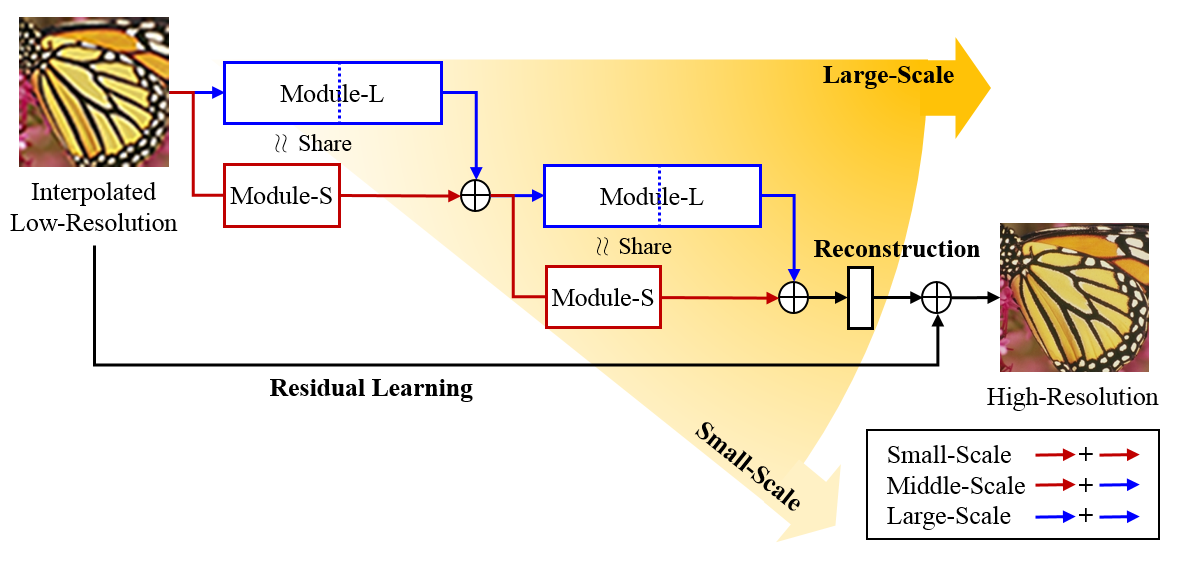}
  \caption{The network architecture of MSSR. We cascade convolutional layers and nonlinear layers (ReLU) repeatedly. An interpolated low-resolution image goes through MSSR and transforms into a high-resolution image. MSSR consists of two convolution modules (Module-L and Module-S), streams of three different scales (Small/Middle/Large-Scale), and a reconstruction module with residual learning.}
  \label{fig:mssr}
\end{figure}

Different kinds of components in an image may be relevant to different scales of neighbourhood. In \cite{msd}, multi-scale neighborhood has been proven effective for super-resolution, which simultaneously integrates local and non-local sparse priors. Multi-scale feature extraction \cite{dehazenet,mscnn} is also effective to represent image patterns. For example, the inception architecture in GoogLeNet \cite{googlenet} uses parallel convolutions with varying filter sizes, and better addresses the issue of aligning objects in input images, resulting in state-of-the-art performance in object recognition. Motivated by this, we propose a multi-scale super-resolution convolutional neural network to improve the performance (see as Fig. \ref{fig:mssr}): low-resolution image is first up-sampled to the desired size by bicubic interpolation, and then MSSR is implemented to predict the detail.

\subsection{Multi-Scale Architecture}

With fixed filter size larger than 1, the receptive field is going larger when network stacks more layers. The proposed architecture is composed of two parallel paths as illustrated in Fig. \ref{fig:mssr}. The upper path (Module-L) stacks $N_L$ convolutional layers which is able to catch a large region of information in the LR image. The other path (Module-S) contains $N_S$ ($N_S<N_L$) convolutional layers to ensure a relatively small receptive filed. The response of the $k$-th convolutional layer in Module-L/S for input $h^k$ is given by
\begin{equation}\label{h}
h^{k+1}=f^{k+1}\left({h^k}\right)=\sigma\left({W^{k+1}*h^{k}+b^{k+1}}\right),
\end{equation}
where $W^{k+1}$ and $b^{k+1}$ are the weights and bias respectively, and $\sigma\left(\cdot\right)$ represents nonlinear operation (ReLU). Here we denote the interpolated low-resolution image as $x$. The output of Module-L is $H_L(x)=f^{N_L}(f^{N_L-1}(...f^{1}(x)))$, and the output of Module-S is $H_S(x)=f^{N_S}(f^{N_S-1}(...f^{1}(x)))$.

For saving consideration, parameters between Module-S and the front part of Module-L are shared. Outputs of the two modules are fused into one, which can take various functional forms (e.g. connection, weighting, and summation). We find that simply summation is efficient enough for our purpose, and the fusion result is generated as $H_f (x)=H_L (x)+H_S (x)$. To further vary the spatial scales of the ensemble architecture, a similar subnetwork is cascaded to the previous one as $F\left(x\right)=H_f(H_f(x))$. A final reconstruction module with $N_r$ convolutional layers is employed to make the prediction. Following \cite{vgg}, size of all convolutional kernels is set to $3 \times 3$ with zero-padding. With respect to the local information involved in LR image, there are streams of three scales (Small/Middle/Large-Scale) corresponding to $2 \times (N_S+N_S+N_r) + 1$, $2 \times (N_S+N_L+N_r) + 1$ and $2 \times (N_L+N_L+N_r) + 1$, respectively. Each layer consists of 64 filters except for the last reconstruction layer, which contains only one single filter without nonlinear operation.

\subsection{Multi-Scale Residual Learning}

High-frequency content is more important for HR restoration, such as gradient features taken into account in \cite{set5,bpres,embed}. Since the input is highly similar to the output in super-resolution problem, the proposed network (MSSR) focuses on high-frequency details estimation through multi-scale residual learning.

The given training set $\{ {x_s^{\left( i \right)},{y^{\left( i \right)}}} \}_{\{i,s\} = \{1,1\}}^{\{N,S\}}$ includes $N$ pairs of multi-scale LR images   $x_s^{\left( i \right)}$ with $S$ scale factors and HR image ${y^{\left( i \right)}}$. Multi-scale residual image for each sample is computed as $r_s^{\left( i \right)} = {y^{\left( i \right)}} - x_s^{\left( i \right)}$. The goal of MSSR is to learn the nonlinear mapping $F\left( x \right)$ from multi-scale LR images $x_s^{\left( i \right)}$ to predict the residual image $r_s^{\left( i \right)}$. The network parameters $\Theta  = \left\{ {{W^k},{b^k}} \right\}$ are achieved through minimizing the loss function as
\begin{equation}\label{f}
\begin{array}{l}
L\left( \Theta  \right) = \dfrac{1}{{2NS}}\sum\limits_{i = 1}^N {\sum\limits_{s = 1}^S {{{\left\| {r_s^{\left( i \right)} - F\left( {x_s^{\left( i \right)};\Theta } \right)} \right\|}^2}} } \\
 = \dfrac{1}{{2NS}}\sum\limits_{i = 1}^N {\sum\limits_{s = 1}^S {{{\left\| {{y^{\left( i \right)}} - \left( {x_s^{\left( i \right)} + F\left( {x_s^{\left( i \right)};\Theta } \right)} \right)} \right\|}^2}} }
\end{array}
\end{equation}

With multi-scale residual learning, we only train a general model for multiple up-scale factors. For LR images $x_s^{\left( i \right)}$ with different down sampling scales $s$, even the same region size in LR images may contain different information content. In the work of Dong et al. \cite{fsrcnn}, a small patch in LR space could cover almost all information of a large patch in HR. For multiple up-scale samples, a model with only one single receptive field cannot make the best of them all simultaneously. However, our multi-scale network is capable of handling this problem. The advantages of multi-scale learning include not only memory and time saving, but also a way to adapt the model for different down sampling scales.

\section{Experiments}

\subsection{Datasets}

\textbf{Training dataset.} The model is trained on 91 images from Yang et al. \cite{srsr} and 200 images from the training set of Berkeley Segmentation Dataset (BSD) \cite{b100}, which are widely used for super-resolution problem \cite{srcnn2,vdsr,fsrcnn,RFL}. As in \cite{fsrcnn}, to make full use of the training data, we apply data augmentation in two ways: 1) Rotate the images with the degree of $90^{\circ}$, $180^{\circ}$ and $270^{\circ}$. 2) Downscale the images with the factor of 0.9, 0.8, 0.7 and 0.6. Following the sample cropping in \cite{vdsr}, training images are cropped into sub-images of size $41 \times 41$ with non-overlapping. In addition, to train a general model for multiple up-scale factors, we combine LR-HR pairs of three up-scale size ($\times 2, \times 3, \times 4$) into one.

\textbf{Test dataset.} The proposed method is evaluated on four publicly available benchmark datasets: Set5 \cite{set5} and Set14 \cite{set14} provide 5 and 14 images respectively; B100 \cite{b100} contains 100 natural images collected from BSD; Urban100 \cite{selfex} consists of 100 high-resolution images rich of structures in real-world. Following previous works \cite{selfex,fsrcnn,vdsr}, we transform the images to YCbCr color space and only apply the algorithm on the luminance channel, since human vision is more sensitive to details in intensity than in color.

\subsection{Experimental Settings}

In the experiments, the \emph{Caffe} \cite{caffe} package is implemented to train the proposed MSSR with Adam \cite{adam}. To ensure varying receptive field scales, we set $N_L=9$, $N_S=2$ and $N_r=2$ respectively. That is, each Module-L in Fig. \ref{fig:mssr} stacks 9 convolutional layers, while Module-S stacks 2 layers. The reconstruction module is built of 2 layers. Thus, the longest path in the network consists of 20 convolutional layers totally, and there are streams of three different scales corresponding to 13, 27 and 41. Model weights are initialized according to the approach described in \cite{msra}. Learning rate is initially set to $10^{-4}$ and decreases by the factor of 10 after 80 epochs. Training phase stops at 100 epochs. We set the parameters of batch-size, momentum and weight decay to 64, 0.9 and $10^{-4}$ respectively.

\subsection{Results}
\begin{figure}[ht]
\centering
\includegraphics[width=0.16\linewidth]{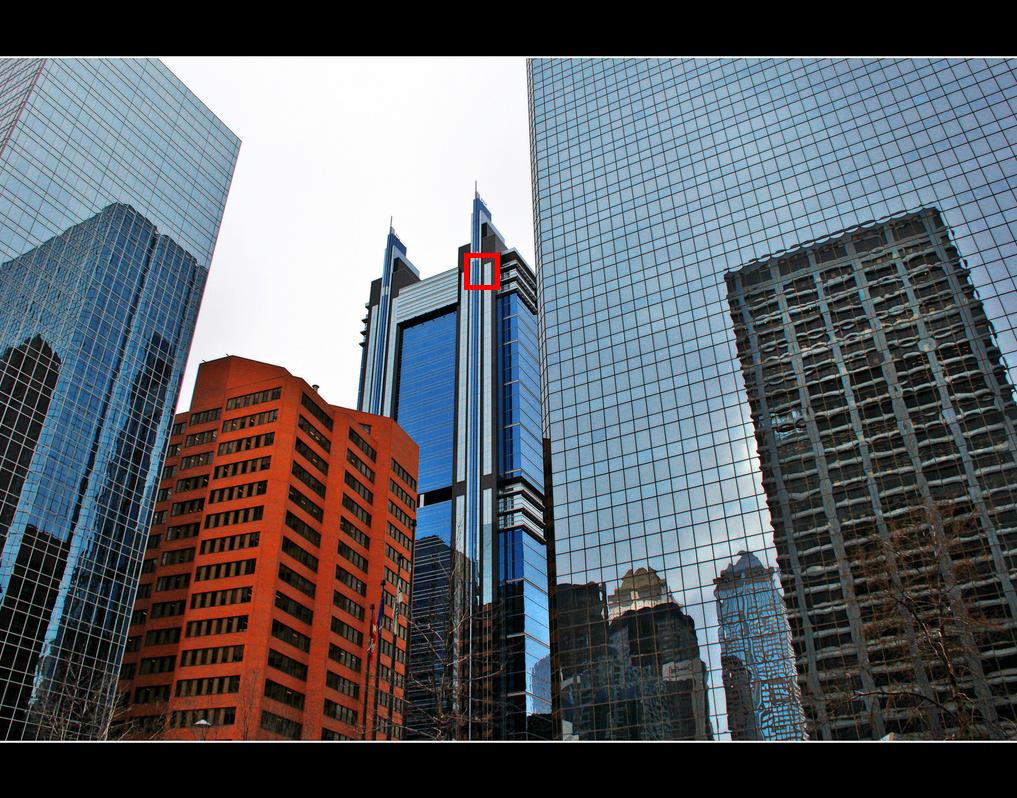}
\includegraphics[width=0.16\linewidth]{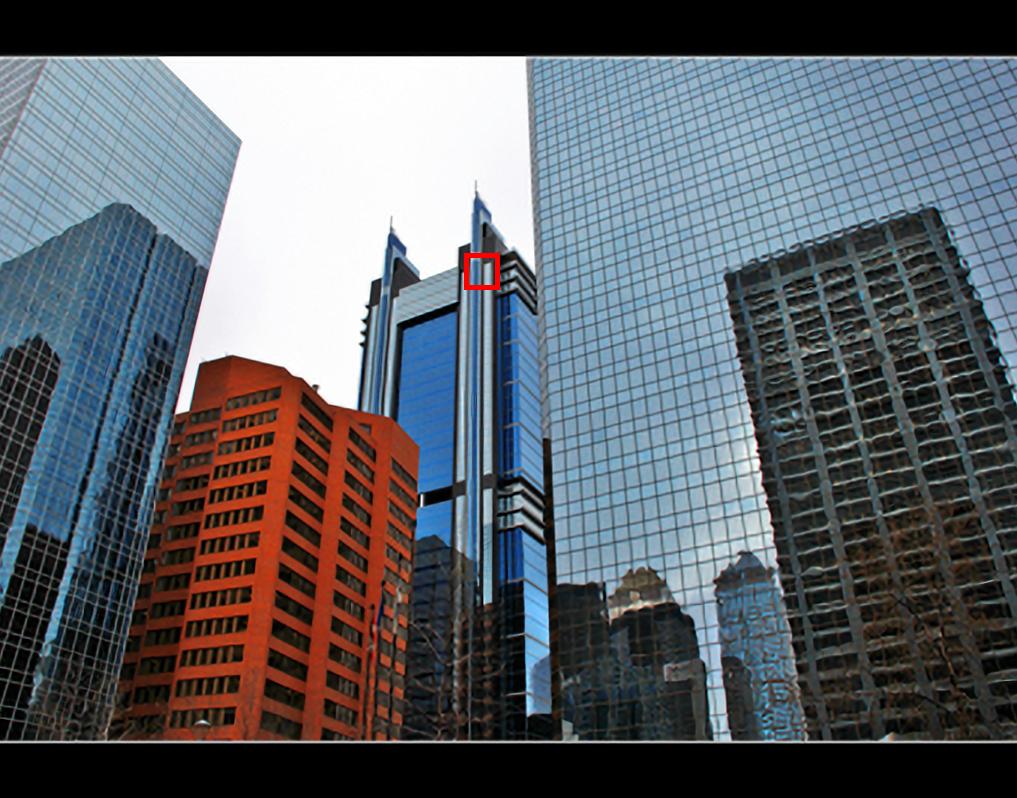}
\includegraphics[width=0.16\linewidth]{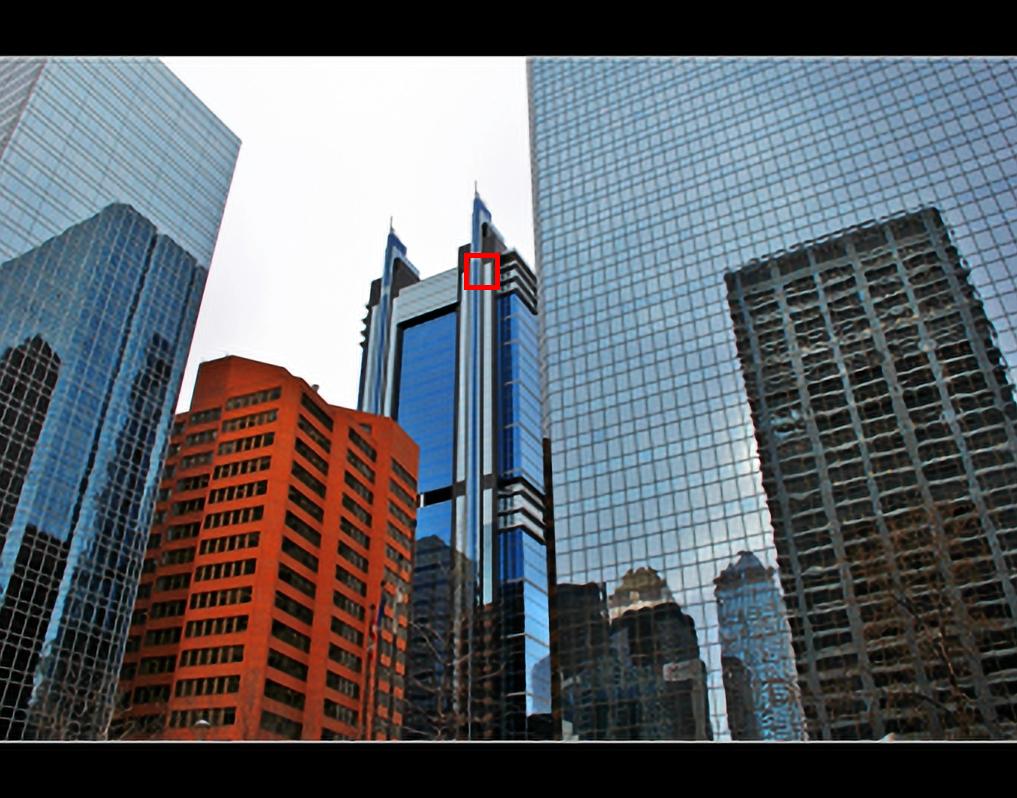}
\includegraphics[width=0.16\linewidth]{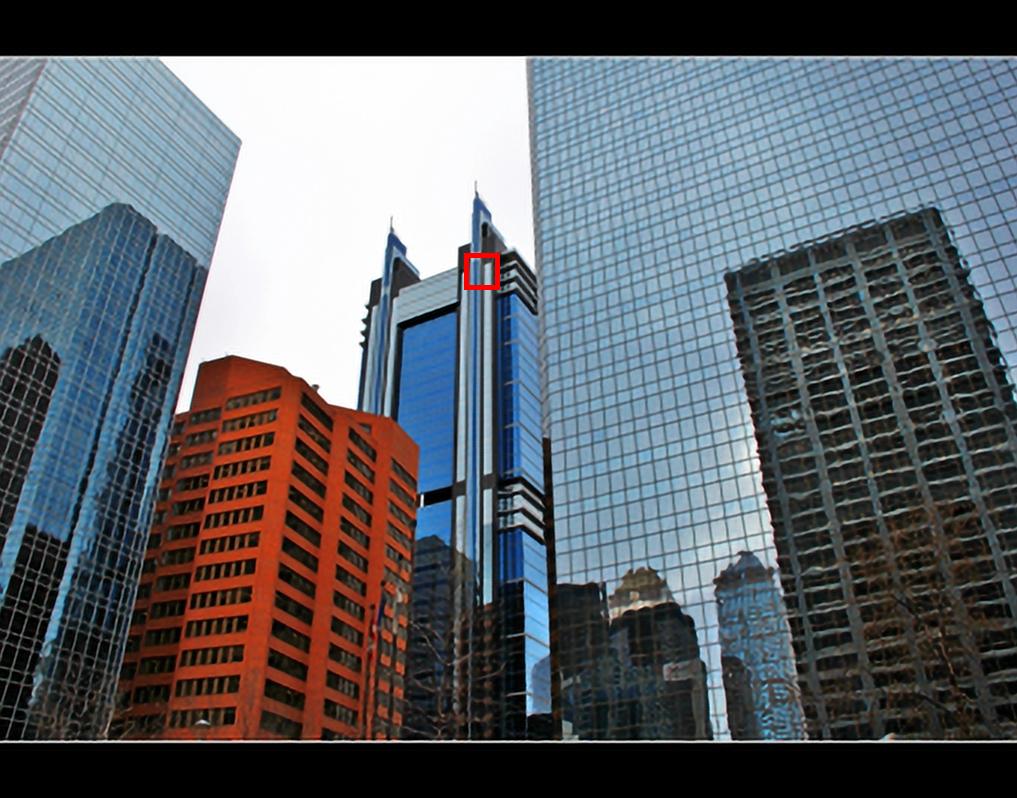}
\includegraphics[width=0.16\linewidth]{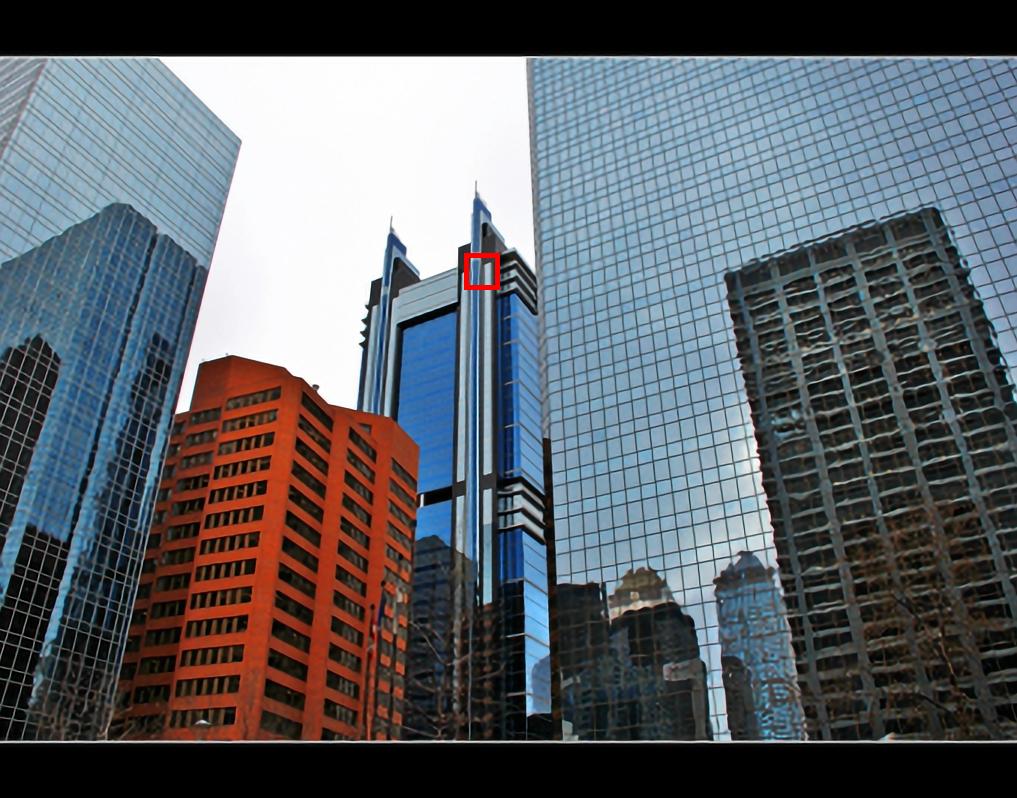}
\includegraphics[width=0.16\linewidth]{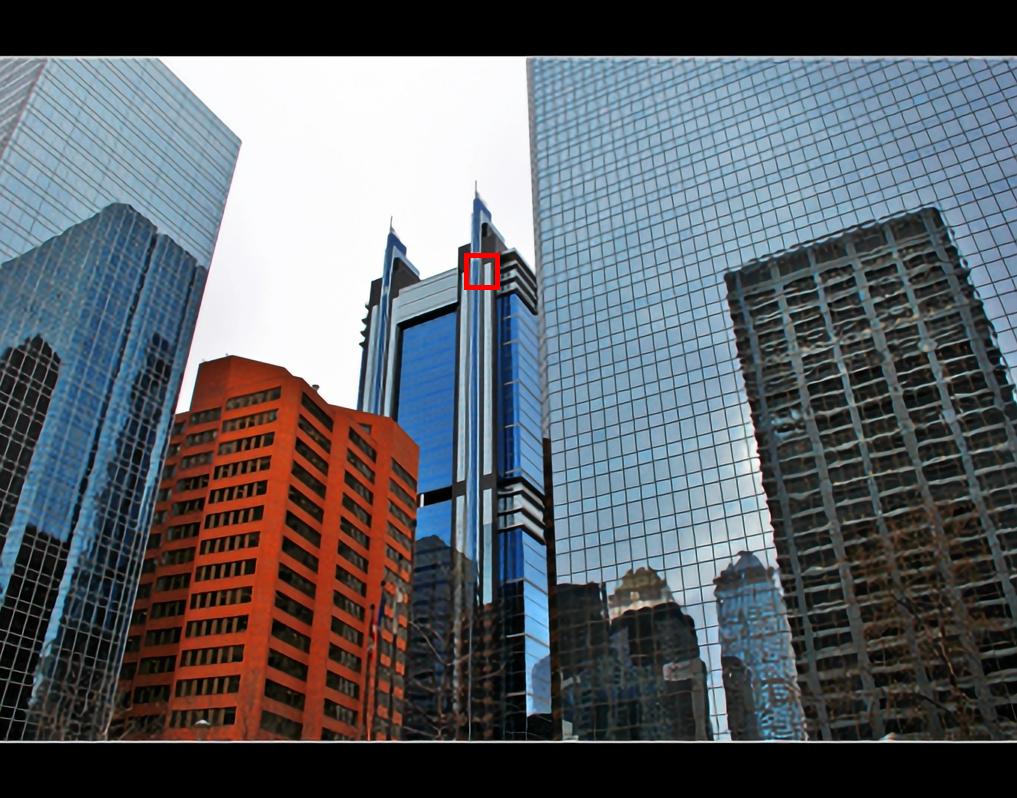}\\
\includegraphics[width=0.16\linewidth]{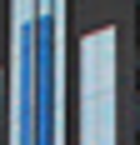}
\includegraphics[width=0.16\linewidth]{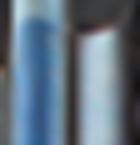}
\includegraphics[width=0.16\linewidth]{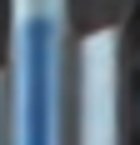}
\includegraphics[width=0.16\linewidth]{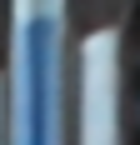}
\includegraphics[width=0.16\linewidth]{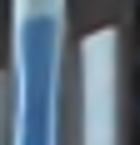}
\includegraphics[width=0.16\linewidth]{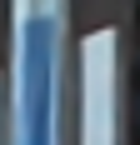}\\
\begin{tabular}{C{1.9cm}C{1.9cm}C{1.9cm}C{1.9cm}C{2.1cm}c}
Original&SelfEx \cite{selfex}&SRCNN \cite{srcnn2}&FSRCNN \cite{fsrcnn}&VDSR \cite{vdsr}&MSSR\\
PSNR/SSIM&25.69/0.7940&25.48/0.7712&25.63/0.7815&\textcolor{blue}{26.58}/\textcolor{blue}{0.8285}&\textcolor{red}{27.01}/\textcolor{red}{0.8391}
\end{tabular}
\caption{Super-resolution results of \emph{img099} (Urban100) with scale factor x3. Line is straightened and sharpened in MSSR, whereas the other methods give blurry or distorted lines.}
\label{fig:set14}
\end{figure}

\begin{figure}[ht]
\centering
\includegraphics[width=0.16\linewidth]{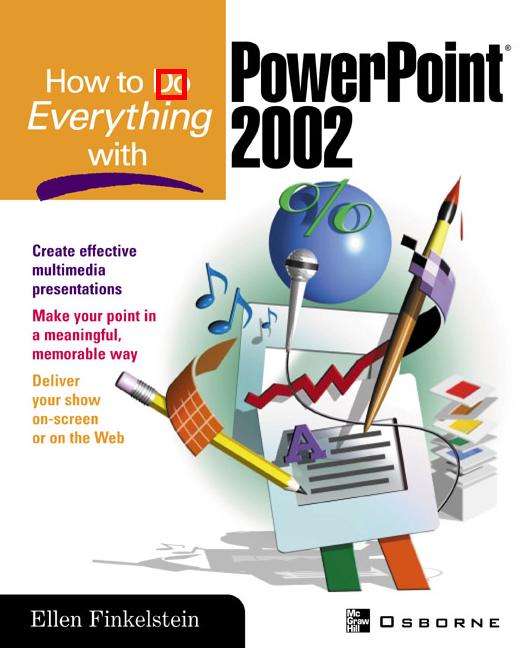}
\includegraphics[width=0.16\linewidth]{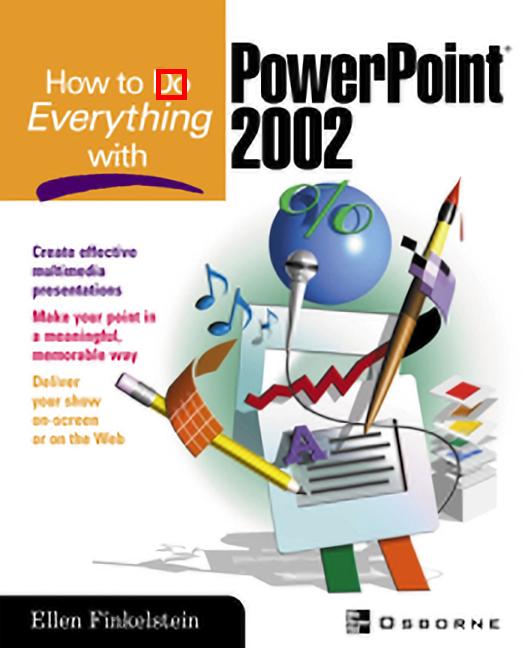}
\includegraphics[width=0.16\linewidth]{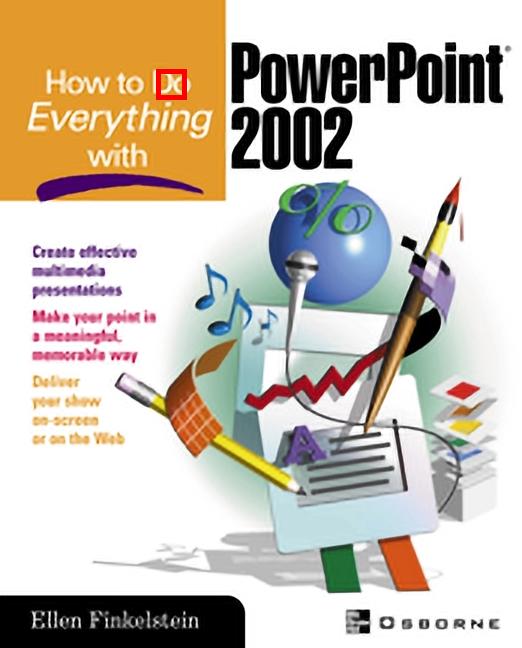}
\includegraphics[width=0.16\linewidth]{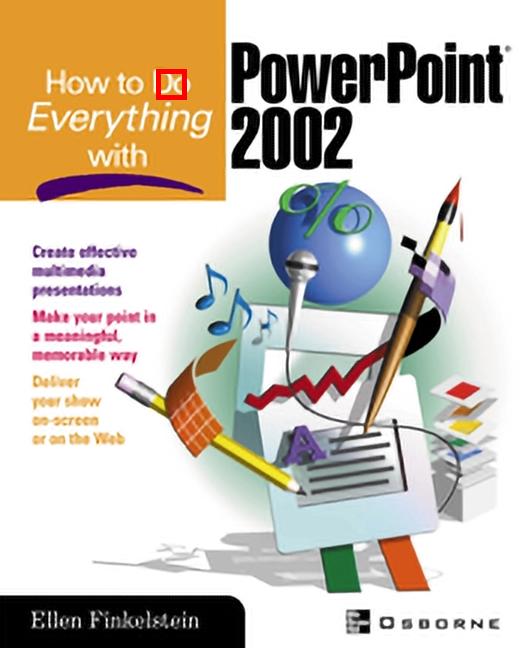}
\includegraphics[width=0.16\linewidth]{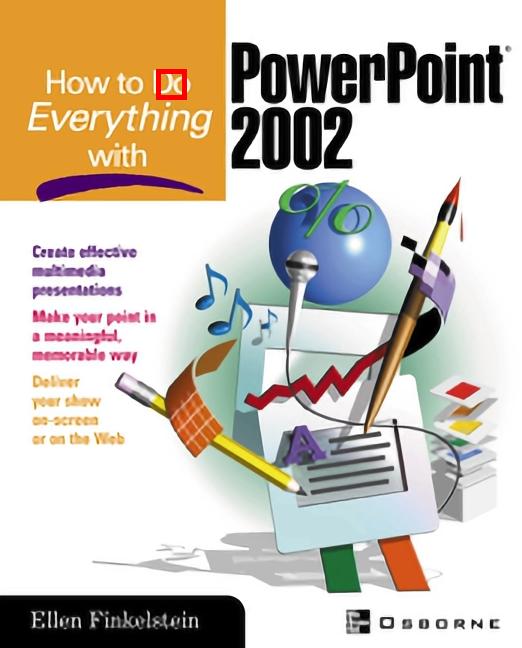}
\includegraphics[width=0.16\linewidth]{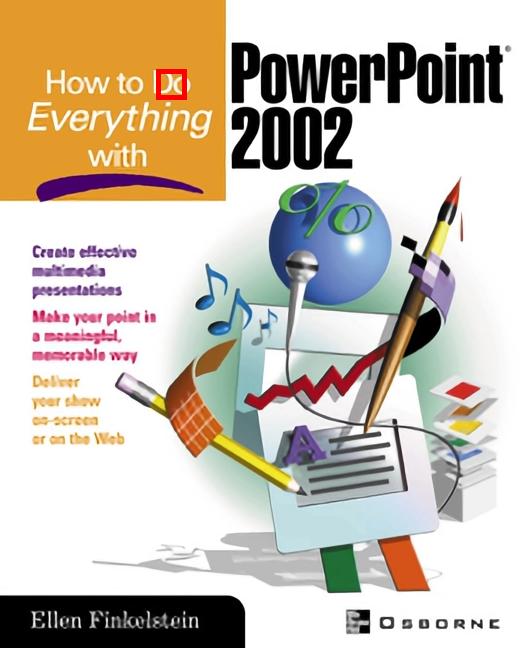}\\
\includegraphics[width=0.16\linewidth]{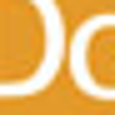}
\includegraphics[width=0.16\linewidth]{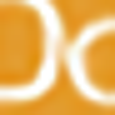}
\includegraphics[width=0.16\linewidth]{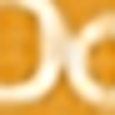}
\includegraphics[width=0.16\linewidth]{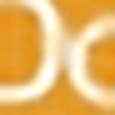}
\includegraphics[width=0.16\linewidth]{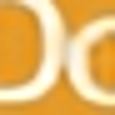}
\includegraphics[width=0.16\linewidth]{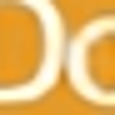}
\begin{tabular}{C{1.9cm}C{1.9cm}C{1.9cm}C{1.9cm}C{2.1cm}c}
Original&SelfEx \cite{selfex}&SRCNN \cite{srcnn2}&FSRCNN \cite{fsrcnn}&VDSR \cite{vdsr}&MSSR\\
PSNR/SSIM&27.08/0.9481&27.04/0.9393&27.11/0.9417&\textcolor{blue}{27.86}/\textcolor{blue}{0.9616}&\textcolor{red}{28.48}/\textcolor{red}{0.9674}
\end{tabular}
\caption{Super-resolution results of \emph{ppt3} (Set14) with scale factor x3. Texts in MSSR are sharp and legible, while character edges are blurry in the other methods.}
\label{fig:ppt}
\end{figure}

To quantitatively assess the proposed model, MSSR is evaluated for three different up-scale factors from 2 to 4 on four testing datasets aforementioned. We compute the Peak Signal-to-Noise Ratio (PSNR) and structural similarity (SSIM) of the results to compare with some recent competitive methods, including A+ \cite{ap}, SelfEx \cite{selfex}, SRCNN \cite{srcnn2}, FSRCNN \cite{fsrcnn} and VDSR \cite{vdsr}. As shown in Table \ref{tab:result}, we can see that the proposed MSSR outperforms other methods almost on every up-scale factor and each test set. The only suboptimal result is the PSNR on B100 of up-scale factor 4, which is slightly lower than VDSR \cite{vdsr}, but still competitive with a higher SSIM. Visual comparisons can be found in Fig. \ref{fig:set14} and Fig. \ref{fig:ppt}.


\begin{figure}[!h]
\centering
\includegraphics[width=0.6\linewidth]{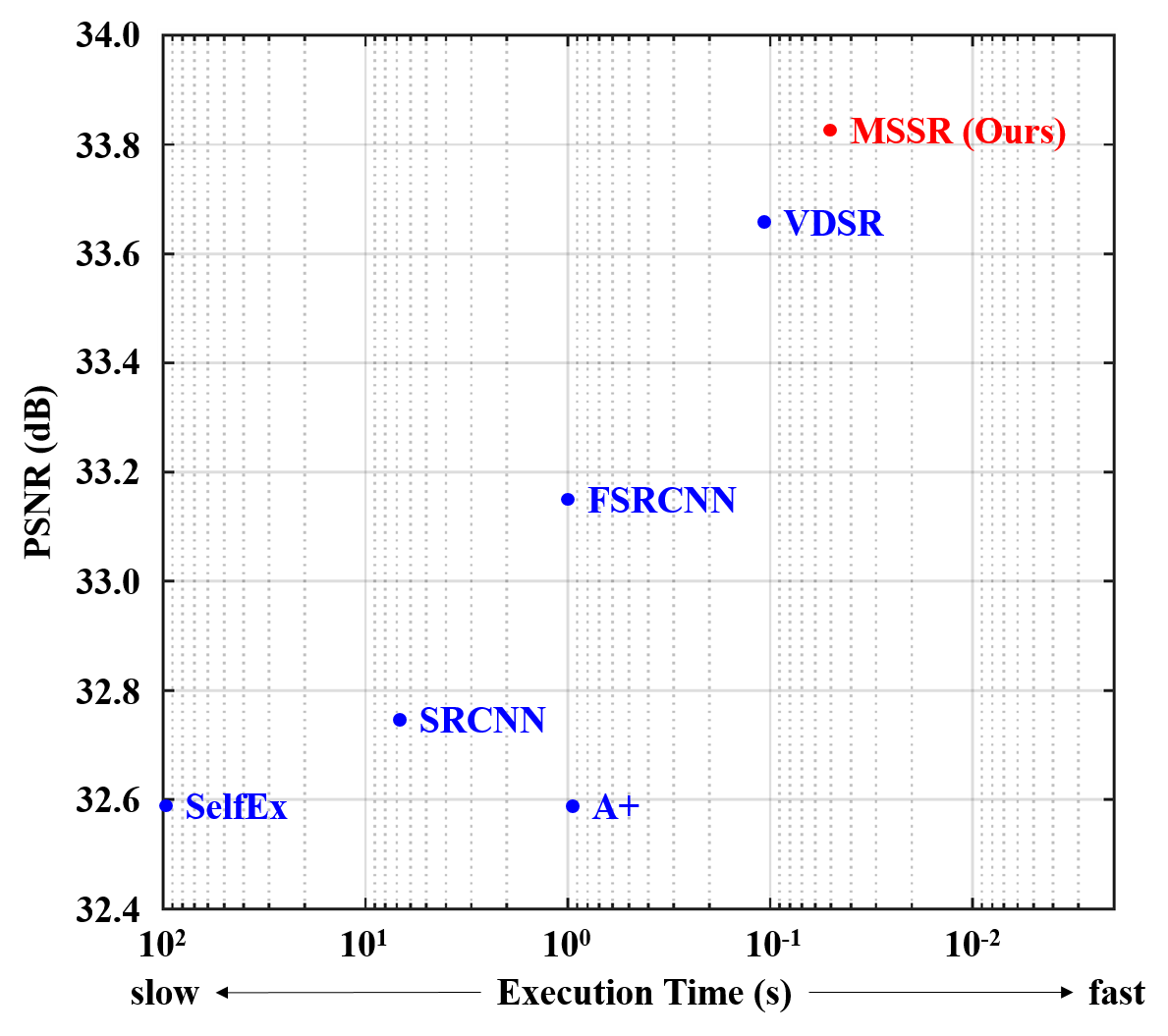}
\caption{Our MSSR achieves more accurate and efficient results for scale factor x3 on dataset Set5 in comparison to the state-of-the-art methods.}
\label{fig:speed}
\end{figure}

As for effectiveness, we evaluate the execution time using the public code of state-of-the-art methods. The experiments are conducted with an Intel CPU (Xeon E5-2620, 2.1 GHz) and an NVIDIA GPU (GeForce GTX 1080). Fig. \ref{fig:speed} shows the PSNR performance of several state-of-the-art methods for super-resolution versus the execution time. The proposed MSSR network achieves better super-resolution quality than existing methods, and are tens of times faster.

\begin{table}
\center
\caption{Average PSNR/SSIM for scale factors x2, x3 and x4 on datasets Set5 \cite{set5}, Set14 \cite{set14}, B100 \cite{b100} and Urban100 \cite{selfex}. \textcolor{red}{Red} color indicates the best performance and \textcolor{blue}{blue} color indicates the second best performance. \protect\footnotemark[1]}
\scriptsize
\begin{tabular}{|c|c|c|c|c|c|c|c|}
\hline
Dataset&Scale&A+ \cite{ap}&SelfEx \cite{selfex}&SRCNN \cite{srcnn2}&FSRCNN \cite{fsrcnn} &VDSR \cite{vdsr}& MSSR\\
\hline
\hline
\multirow{3}{*}{Set5}&x2&36.54/0.9544&36.49/0.9537&36.66/0.9542&37.00/0.9558&\textcolor{blue}{37.53}/\textcolor{blue}{0.9587}&\textcolor{red}{37.62}/\textcolor{red}{0.9592} \\
&x3&32.58/0.9088&32.58/0.9093&32.75/0.9090&33.16/0.9140&\textcolor{blue}{33.66}/\textcolor{blue}{0.9213}&\textcolor{red}{33.82}/\textcolor{red}{0.9226} \\
&x4&30.28/0.8603&30.31/0.8619&30.49/0.8628&30.71/0.8657&\textcolor{blue}{31.35}/\textcolor{blue}{0.8838}&\textcolor{red}{31.42}/\textcolor{red}{0.8849} \\
\hline
\hline
\multirow{3}{*}{Set14}&x2&32.28/0.9056&32.22/0.9034&32.45/0.9067&32.63/0.9088&\textcolor{blue}{33.03}/\textcolor{blue}{0.9124}&\textcolor{red}{33.11}/\textcolor{red}{0.9133} \\
&x3&29.13/0.8188&29.16/0.8196&29.30/0.8215&29.43/0.8242&\textcolor{blue}{29.77}/\textcolor{blue}{0.8314}&\textcolor{red}{29.86}/\textcolor{red}{0.8332} \\
&x4&27.32/0.7491&27.40/0.7518&27.50/0.7513&27.59/0.7535&\textcolor{blue}{28.01}/\textcolor{blue}{0.7674}&\textcolor{red}{28.05}/\textcolor{red}{0.7686} \\
\hline
\hline
\multirow{3}{*}{B100}&x2&31.21/0.8863&31.18/0.8855&31.36/0.8879&31.50/0.8906&\textcolor{blue}{31.90}/\textcolor{blue}{0.8960}&\textcolor{red}{31.94}/\textcolor{red}{0.8966} \\
&x3&28.29/0.7835&28.29/0.7840&28.41/0.7863&28.52/0.7893&\textcolor{blue}{28.82}/\textcolor{blue}{0.7976}&\textcolor{red}{28.85}/\textcolor{red}{0.7985} \\
&x4&26.82/0.7087&26.84/0.7106&26.90/0.7103&26.96/0.7128&\textcolor{red}{27.29}/\textcolor{blue}{0.7251}&\textcolor{blue}{27.28}/\textcolor{red}{0.7256} \\
\hline
\hline
\multirow{3}{*}{Urban100}&x2&29.20/0.8938&29.54/0.8967&29.51/0.8946&29.85/0.9009&\textcolor{blue}{30.76}/\textcolor{blue}{0.9140}&\textcolor{red}{30.84}/\textcolor{red}{0.9149} \\
&x3&26.03/0.7973&26.44/0.8088&26.24/0.7991&26.42/0.8064&\textcolor{blue}{27.14}/\textcolor{blue}{0.8279}&\textcolor{red}{27.20}/\textcolor{red}{0.8295} \\
&x4&24.32/0.7183&24.79/0.7374&24.52/0.7226&24.60/0.7258&\textcolor{blue}{25.18}/\textcolor{blue}{0.7524}&\textcolor{red}{25.19}/\textcolor{red}{0.7535} \\
\hline
\end{tabular}
\label{tab:result}
\end{table}

\footnotetext[1]{All the output images are cropped to the same size as SRCNN \cite{srcnn2} for fair comparisons.}

\section{Conclusion}
In this paper, we highlight the importance of scales in super-resolution problem, which is neglected in the previous work. Instead of simply enlarge the size of input patches, we proposed a multi-scale convolutional neural network for single image super-resolution. Combining paths of different scales enables the model to synthesize a wider range of receptive fields. Since different components in images may be relevant to a diversity of neighbor sizes, the proposed network can benefit from multi-scale features. Our model generalizes well across different up-scale factors. Experimental results reveal that our approach can achieve state-of-the-art results on standard benchmarks with a relatively high speed.

\bibliographystyle{splncs03}
\bibliography{refs}

\begin{thebibliography}{10}
\providecommand{\url}[1]{\texttt{#1}}
\providecommand{\urlprefix}{URL }

\bibitem{set5}
Bevilacqua, M., Roumy, A., Guillemot, C., Alberi-Morel, M.L.: Low-complexity
  single-image super-resolution based on nonnegative neighbor embedding  (2012)

\bibitem{bpres}
Bevilacqua, M., Roumy, A., Guillemot, C., Morel, M.L.A.: Super-resolution using
  neighbor embedding of back-projection residuals. In: Digital Signal
  Processing (DSP), 2013 18th International Conference on. pp. 1--8. IEEE
  (2013)

\bibitem{dehazenet}
Cai, B., Xu, X., Jia, K., Qing, C., Tao, D.: Dehazenet: An end-to-end system
  for single image haze removal. IEEE Transactions on Image Processing  25(11),
   5187--5198 (2016)

\bibitem{embed}
Chang, H., Yeung, D.Y., Xiong, Y.: Super-resolution through neighbor embedding.
  In: Computer Vision and Pattern Recognition, 2004. CVPR 2004. Proceedings of
  the 2004 IEEE Computer Society Conference on. vol.~1, pp. I--I. IEEE (2004)

\bibitem{bicubic}
De~Boor, C.: Bicubic spline interpolation. Studies in Applied Mathematics
  41(1-4),  212--218 (1962)

\bibitem{srcnn1}
Dong, C., Loy, C.C., He, K., Tang, X.: Learning a deep convolutional network
  for image super-resolution. In: European Conference on Computer Vision. pp.
  184--199. Springer (2014)

\bibitem{srcnn2}
Dong, C., Loy, C.C., He, K., Tang, X.: Image super-resolution using deep
  convolutional networks. IEEE transactions on pattern analysis and machine
  intelligence  38(2),  295--307 (2016)

\bibitem{fsrcnn}
Dong, C., Loy, C.C., Tang, X.: Accelerating the super-resolution convolutional
  neural network. In: European Conference on Computer Vision. pp. 391--407.
  Springer (2016)

\bibitem{lanczos}
Duchon, C.E.: Lanczos filtering in one and two dimensions. Journal of Applied
  Meteorology  18(8),  1016--1022 (1979)

\bibitem{rcnn}
Girshick, R., Donahue, J., Darrell, T., Malik, J.: Rich feature hierarchies for
  accurate object detection and semantic segmentation. In: Proceedings of the
  IEEE conference on computer vision and pattern recognition. pp. 580--587
  (2014)

\bibitem{msra}
He, K., Zhang, X., Ren, S., Sun, J.: Delving deep into rectifiers: Surpassing
  human-level performance on imagenet classification. In: Proceedings of the
  IEEE international conference on computer vision. pp. 1026--1034 (2015)

\bibitem{selfex}
Huang, J.B., Singh, A., Ahuja, N.: Single image super-resolution from
  transformed self-exemplars. In: Proceedings of the IEEE Conference on
  Computer Vision and Pattern Recognition. pp. 5197--5206 (2015)

\bibitem{caffe}
Jia, Y., Shelhamer, E., Donahue, J., Karayev, S., Long, J., Girshick, R.,
  Guadarrama, S., Darrell, T.: Caffe: Convolutional architecture for fast
  feature embedding. In: Proceedings of the 22nd ACM international conference
  on Multimedia. pp. 675--678. ACM (2014)

\bibitem{vdsr}
Kim, J., Kwon~Lee, J., Mu~Lee, K.: Accurate image super-resolution using very
  deep convolutional networks. In: Proceedings of the IEEE Conference on
  Computer Vision and Pattern Recognition. pp. 1646--1654 (2016)

\bibitem{prior}
Kim, K.I., Kwon, Y.: Single-image super-resolution using sparse regression and
  natural image prior. IEEE transactions on pattern analysis and machine
  intelligence  32(6),  1127--1133 (2010)

\bibitem{adam}
Kingma, D., Ba, J.: Adam: A method for stochastic optimization. arXiv preprint
  arXiv:1412.6980  (2014)

\bibitem{b100}
Martin, D., Fowlkes, C., Tal, D., Malik, J.: A database of human segmented
  natural images and its application to evaluating segmentation algorithms and
  measuring ecological statistics. In: Computer Vision, 2001. ICCV 2001.
  Proceedings. Eighth IEEE International Conference on. vol.~2, pp. 416--423.
  IEEE (2001)

\bibitem{RFL}
Schulter, S., Leistner, C., Bischof, H.: Fast and accurate image upscaling with
  super-resolution forests. In: Proceedings of the IEEE Conference on Computer
  Vision and Pattern Recognition. pp. 3791--3799 (2015)

\bibitem{subpixel}
Shi, W., Caballero, J., Husz{\'a}r, F., Totz, J., Aitken, A.P., Bishop, R.,
  Rueckert, D., Wang, Z.: Real-time single image and video super-resolution
  using an efficient sub-pixel convolutional neural network. In: Proceedings of
  the IEEE Conference on Computer Vision and Pattern Recognition. pp.
  1874--1883 (2016)

\bibitem{vgg}
Simonyan, K., Zisserman, A.: Very deep convolutional networks for large-scale
  image recognition. arXiv preprint arXiv:1409.1556  (2014)

\bibitem{googlenet}
Szegedy, C., Liu, W., Jia, Y., Sermanet, P., Reed, S., Anguelov, D., Erhan, D.,
  Vanhoucke, V., Rabinovich, A.: Going deeper with convolutions. In:
  Proceedings of the IEEE Conference on Computer Vision and Pattern
  Recognition. pp. 1--9 (2015)

\bibitem{ap}
Timofte, R., De~Smet, V., Van~Gool, L.: A+: Adjusted anchored neighborhood
  regression for fast super-resolution. In: Asian Conference on Computer
  Vision. pp. 111--126. Springer (2014)

\bibitem{srsr}
Yang, J., Wright, J., Huang, T.S., Ma, Y.: Image super-resolution via sparse
  representation. IEEE transactions on image processing  19(11),  2861--2873
  (2010)

\bibitem{mscnn}
Zeng, L., Xu, X., Cai, B., Qiu, S., Zhang, T.: Multi-scale convolutional neural
  networks for crowd counting. arXiv preprint arXiv:1702.02359  (2017)

\bibitem{set14}
Zeyde, R., Elad, M., Protter, M.: On single image scale-up using
  sparse-representations. In: International conference on curves and surfaces.
  pp. 711--730. Springer (2010)

\bibitem{msd}
Zhang, K., Gao, X., Tao, D., Li, X.: Multi-scale dictionary for single image
  super-resolution. In: Computer Vision and Pattern Recognition (CVPR), 2012
  IEEE Conference on. pp. 1114--1121. IEEE (2012)

\end{thebibliography}
\end{document}